\documentclass[conference,a4paper]{IEEEtran}
\IEEEoverridecommandlockouts
\usepackage{cite}
\usepackage[left=1.62cm,right=1.62cm,top=1.9cm]{geometry}
\usepackage{amsmath,amssymb,amsfonts}
\usepackage{algorithm,algcompatible}
\usepackage{algpseudocode}
\usepackage[table]{xcolor}
\usepackage{dblfloatfix}
\usepackage{graphicx}
\usepackage{makecell}
\usepackage{textcomp}
\usepackage{xcolor}
\usepackage{comment}
\usepackage{cuted}
\usepackage{multirow}
\usepackage{changepage}
\usepackage{float}
\restylefloat{figure}%
\usepackage[mathcal]{eucal}
\def\BibTeX{{\rm B\kern-.05em{\sc i\kern-.025em b}\kern-.08em
    T\kern-.1667em\lower.7ex\hbox{E}\kern-.125emX}}
\begin{document}

\title{Variational Density Propagation \\Continual Learning\\
\thanks{This work was supported by the National Science Foundation Awards NSF ECCS-1903466 and NSF OAC- 2234836. We are also grateful to UK EPSRC support through EP/T013265/1 project NSF-EPSRC: ShiRAS. Towards Safe and Reliable Autonomy in Sensor Driven Systems.}
}
\author{\IEEEauthorblockN{Christopher F. Angelini}
\IEEEauthorblockA{\textit{Dept. of Electrical and}\\ \textit{Computer Engineering} \\
\textit{Rowan University}\\
Glassboro, New Jersey, USA \\
angelinic0@rowan.edu}
\and
\IEEEauthorblockN{Nidhal C. Bouaynaya}
\IEEEauthorblockA{\textit{Dept. of Electrical and }\\ \textit{Computer Engineering} \\
\textit{Rowan University}\\
Glassboro, New Jersey, USA \\
bouaynaya@rowan.edu}
\and
\IEEEauthorblockN{Ghulam Rasool}
\IEEEauthorblockA{\textit{Depts. of Machine Learning}\\ \textit{and Neuro-Oncology} \\
\textit{Moffitt Cancer Center \& Research Institute}\\
Tampa, Florida, USA \\
Ghulam.Rasool@moffitt.org}
}

\maketitle

\begin{abstract}
Deep Neural Networks (DNNs) deployed to the real world are regularly subject to out-of-distribution (OoD) data, various types of noise, and shifting conceptual objectives. This paper proposes a framework for adapting to data distribution drift modeled by benchmark Continual Learning datasets. We develop and evaluate a method of Continual Learning that leverages uncertainty quantification from Bayesian Inference to mitigate catastrophic forgetting. We expand on previous approaches by removing the need for Monte Carlo sampling of the model weights to sample the predictive distribution. We optimize a closed-form Evidence Lower Bound (ELBO) objective approximating the predictive distribution by propagating the first two moments of a distribution, i.e. mean and covariance, through all network layers. Catastrophic forgetting is mitigated by using the closed-form ELBO to approximate the Minimum Description Length (MDL) Principle, inherently penalizing changes in the model likelihood by minimizing the KL Divergence between the variational posterior for the current task and the previous task's variational posterior acting as the prior. Leveraging the approximation of the MDL principle, we aim to initially learn a sparse variational posterior and then minimize additional model complexity learned for subsequent tasks. Our approach is evaluated for the task incremental learning scenario using density propagated versions of fully-connected and convolutional neural networks across multiple sequential benchmark datasets with varying task sequence lengths. Ultimately, this procedure produces a minimally complex network over a series of tasks mitigating catastrophic forgetting.
\end{abstract}


\section{Introduction}
A commonly held assumption in deep learning is a network's training and test data distributions are static and accurately represent its deployed environment.  However, Deep Neural Networks (DNNs) deployed in real-world environments are regularly subject to out-of-distribution (OoD) data, various types of noise, and shifting conceptual objectives \cite{Ditzler2015}. Input patterns not generalized by the network after training may inadvertently trigger features in the network leading to incorrect results with high confidence during deployment. While adapting to data drift can be achieved by retraining a DNN on an entire dataset comprised of the original samples and new drifted samples, there are situations where this may be infeasible or unreasonable due to time, cost, computing, or retained data constraints. Additionally, simply training on the new information from the data drift will result in a phenomenon called \textit{catastrophic forgetting}, causing representations of newly trained information to interfere with or overwrite previous representations, resulting in a drop in performance on previously trained samples \cite{French1999,McCloskey1989}.

Reasonable stand-alone performance in real-world environments requires DNNs to indicate OoD data samples and adapt to these anonymous samples without sacrificing performance on previously trained information. To achieve these requirements, the sequential nature of Bayesian inference naturally lends itself as a mathematical framework to quantify uncertainty and continually adapt to new data without forgetting previous information \cite{He2020}.  Variational methods using the Evidence Lower Bound (ELBO) to approximate the true parameter posterior in Bayesian inference are asymptotically equivalent to the Minimum Description Length principle as the sample size grows to infinity  \cite{Grunwald2008}.  Both approaches regularize optimization by penalizing model complexity, preventing the overfitting of training data. We propose to leverage this concept for Continual Learning (CL) to mitigate catastrophic forgetting by minimizing changes to the network parameters over subsequent tasks. 

While many deep learning variational approximations of Bayesian Inference exist, such as Bayes-by-Backprop (BBB) \cite{Blundell2015}, Monte-Carlo Dropout (MCDrop) \cite{Gal2015}, and have been applied to Continual Learning \cite{Nguyen2017}, most rely on Monte Carlo sampling during training and testing to estimate predictive uncertainty. Uncertainty learned by these methods is only based on a few samples of the variational posterior requiring the reparameterization trick \cite{Kingma2013, Blundell2015} using "noise variables" to decouple the network parameters from the variational distribution for dealing with non-differentiable and/or complex models. A major limitation of these approaches is the significant time and computational cost imposed by sampling the variational posterior, where more samples must be drawn to better estimate the uncertainty about a prediction.

Instead, we propose using the general, model-agnostic framework, Variational Density Propagation (VDP) demonstrated in our previous works \cite{Dera2021, Dera2019}, to quantify the predictive uncertainty and aid in the mitigation of catastrophic forgetting via model complexity regularization. This process removes the need for Monte Carlo sampling of the variational posterior by propagating the first two moments of the variational distribution, i.e. mean and covariance, using a first-order Taylor series approximation. To enhance variance expression in the Variational Density Propagation framework, we impose independence across all network parameters resulting in each random variable parameter having a unique mean and variance. We approximate the propagation of the covariance matrix through each layer as a vectorized diagonal of the full covariance matrix representing the variance of each propagated feature to aid computational efficiency.

By modeling changes in each parameter's variational posteriors as a change in model complexity, changes in parameters can be penalized via the approximation of the Minimum Description Length Principle. This approximation is achieved by minimizing the Kullback-Leibler (KL) divergence between the variational posterior and the network prior inherent to the Evidence Lower Bound objective of Variational Inference. While learning additional information and overwriting already trained representations may not strictly increase model complexity, setting the network prior to the variational posterior from the previously learned task will construe any change to the previous variational posterior as increased model complexity. In this manner, an additional task or a data drift's representation can be learned with minimal changes to the variational posterior from the previous task, preserving representations and mitigating catastrophic forgetting.

To demonstrate this approach, our contributions are as follows:
\begin{itemize}
    \item We develop a fully factorized version of the Variational Density Propagation framework, which uses Taylor-series approximation to propagate the first two moments of the variational distribution,  with an approximation of the propagated covariance matrix for reduced computational requirements.
    \item We convert the approximation of Monte Carlo sampling to the propagation of variational moments to approximate the Minimum Description Length Principle.
    \item We apply our framework to the problem of task incremental learning by imposing a model complexity cost over a series of tasks.
    \item We demonstrate catastrophic forgetting mitigation over task incremental learning with multiple sequential benchmark datasets and compare our results to Monte Carlo sampling-based approaches and baseline performance metrics in the Bayesian and deterministic settings. 
\end{itemize}

\section{Bayesian Deep Learning}
In Deep Bayesian Inference, network parameters, $\boldsymbol{\mathcal{W}}$, are represented as random variables with some prior distribution $\boldsymbol{\mathcal{W}} \sim p(\boldsymbol{\mathcal{W}})$. After observing some training set $\mathcal{D}=\{\mathbf{X}^{(i)},\mathbf{y}^{(i)}\}{\substack{n\quad\\i=1}}$, Bayes' Rule is used to determine the posterior distribution $p(\boldsymbol{\mathcal{W}}|\mathcal{D})$. The predictive distribution, $p(\mathbf{y}|\mathbf{X} ,\mathcal{D})$ is determined by marginalizing the parameters, $\boldsymbol{\mathcal{W}}$, within the posterior distribution. Marginalization can also be performed previously unseen data $\mathbf{\Tilde{X}}$  with corresponding output $\mathbf{\Tilde{y}}$ to perform inference after training, as shown in Equation \ref{eq:pred}.
 \begin{equation}\label{eq:pred}
     p(\mathbf{\Tilde{y}}|\mathbf{\Tilde{X}} ,\mathcal{D}) = \int{p(\mathbf{\Tilde{y}}|\mathbf{\Tilde{X}} ,\boldsymbol{\mathcal{W}})p(\boldsymbol{\mathcal{W}}|\mathcal{D}) \quad d\boldsymbol{\mathcal{W}}}
 \end{equation}
The mean of the predictive distribution represents the network's prediction. The predictive variance represents the statistical uncertainty of the network attached to the same prediction. The predictive variance is construed as confidence in an estimation. 

\subsection{Variational Inference}
Despite having an analytical formulation, applying Bayesian Inference to DNNs is intractable due to the required integration of all network parameters \cite{Blundell2015}. To avoid this issue, Variational Inference (VI) is a common technique used to estimate the posterior distribution of a network by converting the intractable, analytical problem of solving the posterior distribution to an optimization problem. This approach imposes a distribution of variational parameters, $q_\theta(\pmb{\Omega})$, over the network parameters and minimizes the Kullback-Leibler (KL) divergence between the variational distribution and the true posterior distribution, shown in Equation \ref{eq:VI_KL}.

{\small
\begin{equation}\label{eq:VI_KL}
\begin{split}
    \min_\theta \text{KL}[q_\theta(\mathbf{\Omega})||p(\mathbf{\Omega}|\mathcal{D})] &= \min_\theta \int{q_\theta(\mathbf{\Omega})} \ln{\frac{q_\theta(\mathbf{\Omega})}{p(\mathbf{\Omega}|\mathcal{D})}}d\mathbf{\Omega} \\ \min_\theta \text{KL}[q_\theta(\mathbf{\Omega})||p(\mathbf{\Omega}|\mathcal{D})] &= \min_\theta \mathop{\mathbb{E}_{q_\theta(\mathbf{\Omega})}} \left[\ln{\frac{q_\theta(\mathbf{\Omega})}{p(\mathbf{\Omega}|\mathcal{D})}}\right]
\end{split}
\end{equation}
}

However, directly minimizing the KL divergence between the variational parameters and the true posterior is also intractable as determining the true posterior from Bayes' rule still requires the evidence derived from the integration over the product of the likelihood function and the prior distribution $p(\mathcal{D}) = \int p(\mathcal{D} \mid \boldsymbol{\Omega}) p(\boldsymbol{\Omega}) d\boldsymbol{\Omega}$. Instead, VI maximizes an equivalent quantity called the Evidence Lower Bound (ELBO) which can be derived from Equation \ref{eq:VI_KL} using Bayes' Rule to rewrite the true posterior, as shown in Equation \ref{eq:ELBO_der}. The log marginal likelihood of the data, $\ln{p(\mathcal{D})}$, is fixed and does not depend on the variational distribution and can be grouped with the KL divergence between the variational posterior and the true posterior becoming the lower bound on the evidence.
\begingroup\makeatletter\def\f@size{8}\check@mathfonts
\begin{equation}\label{eq:ELBO_der}
\begin{split}
    \text{KL}[q_\theta(\mathbf{\Omega})||p(\mathbf{\Omega}|\mathcal{D})] &= \mathop{\mathbb{E}_{q_\theta(\mathbf{\Omega})}}[ \ln{q_\theta(\mathbf{\Omega})} - \ln{{p(\mathbf{\Omega}|\mathcal{D})}}] \\
    \text{KL}[q_\theta(\mathbf{\Omega})||p(\mathbf{\Omega}|\mathcal{D})] &= \mathop{\mathbb{E}_{q_\theta(\mathbf{\Omega})}}\left[ \ln{q_\theta(\mathbf{\Omega})} - \ln{\frac{p(y|X,\mathbf{\Omega})p(\mathbf{\Omega})}{p(\mathcal{D})}} \right] \\
    \text{KL}[q_\theta(\mathbf{\Omega})||p(\mathbf{\Omega}|\mathcal{D})] &= \mathop{\mathbb{E}_{q_\theta(\mathbf{\Omega})}}\left[ \ln{\frac{q_\theta(\mathbf{\Omega})}{p(\mathbf{\Omega})}}  - \ln{p(y|X,\mathbf{\Omega})} \right] + \ln{p(\mathcal{D})}\\
    -\text{ELBO} &=-\mathop{\mathbb{E}_{q_\theta(\mathbf{\Omega})}}[ \ln{p(y|X,\mathbf{\Omega})}] + \text{KL}_{q_\theta(\mathbf{\Omega})}[q_{\theta}(\mathbf{\Omega})||p(\mathbf{\Omega})] \\
\end{split}
\end{equation}
\endgroup
The optimization objective for Variational Inference is the maximization of the ELBO as shown in Equation (\ref{eq:ELBO}). Maximizing this quantity ultimately maximizes the log-likelihood of data given the network parameters and minimizes the KL divergence between the variational posterior and the prior distribution.
\begingroup\makeatletter\def\f@size{9}\check@mathfonts
\begin{equation}\label{eq:ELBO}
    \phi^*=\text{argmax}\left[\mathop{\mathbb{E}_{q_\theta(\mathbf{\Omega})}}[ \ln{p(y|X,\mathbf{\Omega})}] - \text{KL}_{q_\theta(\mathbf{\Omega})}[q_{\theta}(\mathbf{\Omega})||p(\mathbf{\Omega})] \right] \\
\end{equation}
\endgroup

\subsection{Moments Propagation}
The goal for moments propagation is to produce the mean and the covariance of the predictive distribution from which the network's success or failure can be gauged by relating the predictive distribution's mean to the label weighted by the predictive distribution's variance. To achieve moments propagation, all network operations are replaced by operations of random variables. Each algebraic operation in a deterministic network is replaced by the multiplication of a random variable with a constant, multiplication of two random variables, or approximating the non-linear transformation over random variables using a first-order Taylor-series approximation \cite{Dera2021}. As a result, the first and second moments of the variational distribution, mean and covariance, can be propagated layer by layer through the entire network.

We consider a fully connected model for simplicity of notation and without loss of generality. Let $\boldsymbol{x} \in \mathbb{R}^{n}$ be the input to a layer with mean $\boldsymbol{\mu}_x$ and covariance matrix $\boldsymbol{\Sigma}_x$. We assume that the $j^{th}$ model parameter $\boldsymbol{w}_j$ follows a Normal distribution $\boldsymbol{w}_j\sim\mathcal{N}(\boldsymbol{\mu}_{w_j},\boldsymbol{\sigma}_{w_j}^2)$. 
The model parameters are assumed to be independent from each other and the input. 
The first-order Taylor series approximates the first two moments after a non-linear activation function. 
In the following, we present the derivations pertaining to the various model layers.

\subsubsection{Propagation through the $k^{\text{th}}$ linear layer}
To streamline notation, we will exclude the reference to $k$ in our representation. 
Let $\boldsymbol{z} = {\boldsymbol{W}}^T \boldsymbol{x} + \boldsymbol{b}$, where, for the $k^{\text{th}}$ layer, $\boldsymbol{W} \in \mathbb{R}^{n \times m}$ is a random matrix of weights, $\boldsymbol{b} \in \mathbb{R}^{m}$ is a random vector of biases, and $\boldsymbol{z} \in \mathbb{R}^{m}$ is the resulting random vector. Let $\boldsymbol{W}=[\boldsymbol{w}_1, \cdots, \boldsymbol{w}_m]$, where $\boldsymbol{w}_i$ is the $i^{\text{th}}$ column of $\boldsymbol{W}$, with the mean and covariance of $\boldsymbol{w}_i$ represented as $\boldsymbol{\mu}_{w_i}$ and $\boldsymbol{\Sigma}_{w_{i}}$. It follows that the mean and variance elements, $\boldsymbol{\mu}_{z_i}$ and $\boldsymbol{\sigma}^{2}_{z_i}$, contained within the resulting random vector $\boldsymbol{z}$ can be derived for elements $i=1 \cdots m$ with the matrix-vector multiplication as shown in Equation (\ref{eq:prop}). 

\begin{equation}\label{eq:prop}
    \begin{split}
    {\mu_{z_{i}}} &= \boldsymbol{\mu}_{w_i}^T \boldsymbol{\mu}_{x} + \mu_{b_{i}} \\
    {\sigma^{2}_{z_{i}}} &=  \text{tr}\left(\boldsymbol{\Sigma}_{w_{i}} \boldsymbol{\Sigma}_x\right) +
    {{\boldsymbol{\mu}}^{T}_{x}}  {\boldsymbol{\Sigma}_{w_{i}}} {\boldsymbol{\mu}}_{x} +
    \boldsymbol{\mu}_{w_i}^T  \boldsymbol{\Sigma}_x \boldsymbol{\mu}_{w_i}  + {\sigma^2_{b_{i}}} \\
    \end{split}
\end{equation}

The input data for the first layer of the network $k=0$ is treated as deterministic where, in Equation (\ref{eq:prop}), $ \boldsymbol{\Sigma}_x = 0$. The resulting covariance for ${\sigma^{2}_{z_{i}}}$ only depends on $\boldsymbol{\mu_x}^T\boldsymbol{\Sigma}_{w_i}\boldsymbol{\mu_x} $.

For computational efficiency, moment propagation is reformulated to only propagate the diagonal variance elements of the covariance information through the network. As a result of this further approximation, Equation (\ref{eq:prop}) can be reformulated as Equation (\ref{eq:propVec}) below, where $\mu_{w_{i,h}}$ and $\sigma^2_{w_{i,h}}$ are the $h^{\text{th}}$ element of $\boldsymbol{\mu}_{w_i}$ and the $h^{\text{th}}$ diagonal element of $\boldsymbol{\Sigma}_{w_i}$, respectively. 
\begin{equation}\label{eq:propVec}
    \begin{split}
    {\mu_{z_{i}}} &= \boldsymbol{\mu}_{w_i}^T \boldsymbol{\mu}_x + \boldsymbol{\mu}_{b_i} \\
    {\sigma^{2}_{z_{i}}} &= \sum_{h=1}^{n}{({\sigma_{x_h}^2} \sigma_{w_{i,h}}^2 + \mu_{x_h}^2 \sigma_{w_{i,h}}^2 + \sigma_{x_h}^2 \mu_{w_{i,h}}^2) + \sigma_{b_i}^2}
    \end{split}
\end{equation}

It is important to note that as Equation (\ref{eq:propVec}) is a further approximation, there is a loss in the fidelity of the propagated variances due to subsequent network layers' reliance on the covariance elements of previous layers to compute the true covariance values. Convolution operations are treated similarly, where kernels and underlying image patches are vectorized before moment propagation.

\subsubsection{Propagation through a non-linear activation}
Let $\boldsymbol{g}=\Psi(\boldsymbol{z})$ represent some non-linear activation function (e.g. ReLU, Hyperbolic Tangent, Softmax) of a random vector input $\boldsymbol{z} \in \mathbb{R}^{m}$ with mean $\boldsymbol{\mu}_z$ and covariance $\boldsymbol{\Sigma}_z$. The mean and covariance for the resulting random vector $\boldsymbol{g}$ can be approximated using the first-order Taylor series approximation \cite{Dera2021, Dera2019, Papoulis2002} in Equation (\ref{eq:nonlin}) where  $\odot$ is the element-wise product of the incoming covariance matrix, $\boldsymbol{\Sigma}_z$, and the squared gradient, $\nabla$, of non-linear function with respect to the incoming mean, $\boldsymbol{\mu}_z$.  

\begin{equation}\label{eq:nonlin}
    \begin{split}
    \boldsymbol{\mu}_g &\approx \Psi(\boldsymbol{\mu}_z) \\
    \boldsymbol{\Sigma}_g &\approx \boldsymbol{\Sigma}_z \odot \nabla\Psi(\boldsymbol{\mu}_z)\nabla\Psi(\boldsymbol{\mu}_z)^T
    \end{split}
\end{equation}

Similarly, to reduce computational complexity, the diagonal variance elements are vectorized, and the covariance elements are ignored. The vectorized form of Equation (\ref{eq:nonlin}) is shown in Equation (\ref{eq:nonlinVec}), where $\boldsymbol{\sigma_z^2}$ represents the vectorized form of the diagonal variance elements. In the vectorized form, the outer product used for constructing the correlation matrix in Equation (\ref{eq:nonlin}) can be replaced by the squared gradient of the non-linear activation function with respect to the input mean due to the off-diagonal elements are no longer required for the Hadamard product with the input covariance matrix. 

\begin{equation}\label{eq:nonlinVec}
    \begin{split}
    \boldsymbol{\mu}_g &\approx \Psi(\boldsymbol{\mu_z}) \\
    \boldsymbol{\sigma}_{g_i}^2 &\approx \boldsymbol{\sigma}_{z_i}^2 \odot \left(\frac{\partial \Psi}{\partial z_i}(\boldsymbol{\mu}_z)\right)^2
    \end{split}
\end{equation}

For the Softmax classification layer at the output of the network $\boldsymbol{\mu}_{\hat{y}}$ and $\boldsymbol{\sigma}^2_{\hat{y}}$ are representative of the variational distribution that can be used to infer the ELBO objective function.

\begin{figure*}[t]
\small{
\begin{equation}\label{eq:ELBO_closed}
    \text{ELBO} = -\frac{N}{2}\ln{(2\pi)} - \frac{1}{2}\ln{(|\boldsymbol{\Sigma_{\hat{y}}}|)} - \frac{1}{2}\left({(\boldsymbol{y} - \boldsymbol{\mu_{\hat{y}}})}^T\boldsymbol{\Sigma_{\hat{y}}}^{-1}(\boldsymbol{y} - \boldsymbol{\mu_{\hat{y}}})\right) - \frac{\tau}{2}\sum^{\boldsymbol{|\Omega|}}_{i=1}{\left(-1 + \frac{(\mu_{q_{w_i}} - \mu_{p_i})^2}{\sigma^2_{q_{w_i}}} + \ln\left(\frac{\sigma_{p_i}^2}{\sigma_{q_{w_i}}^2}\right) + \frac{\sigma^2_{q_{w_i}}}{\sigma^2_{p_i}}\right)}
\end{equation}

\begin{equation}\label{eq:ELBO_closed_vec}
    \text{ELBO} = -\frac{N}{2}\ln{(2\pi)} - \frac{1}{2}\sum^N_{n=1}{\ln{(\sigma}^2_{\hat{y}}}_n) + \frac{1}{2}\sum^{N}_{n=1}{\frac{(y_n - \mu_{{\hat{y}}_n})^2}{\sigma^2_{\hat{y}_n}}} - \frac{\tau}{2}\sum^{\boldsymbol{|\Omega|}}_{i=1}{\left(-1 + \frac{(\mu_{q_{w_i}} - \mu_{p_i})^2}{\sigma^2_{q_{w_i}}} - \ln\left(\frac{\sigma_{p_i}^2}{\sigma_{q_{w_i}}^2}\right) - \frac{\sigma^2_{q_{w_i}}}{\sigma^2_{p_i}}\right)}
\end{equation}}
\end{figure*}

\subsection{Closed Form ELBO}
Recalling the objective function for Variational Inference, presented in Equation (\ref{eq:ELBO}), using the propagated values $\boldsymbol{\mu_{\hat{y}}}$ and $\boldsymbol{\sigma^2_{\hat{y}}}$ of the variational distribution $q(\boldsymbol{\Omega})$ the ELBO can be written in closed form, as shown in Equation (\ref{eq:ELBO_closed}). The weighting variable $\boldsymbol{\tau}$ is added to control the level of compression toward the prior induced by the KL divergence term in the ELBO.
The vectorized form of the closed-form ELBO changes the log determinant of the covariance matrix $\boldsymbol{\Sigma_{\hat{y}}}$ to the log sum of all the variational distributions, effectively producing the log product of the diagonal variance values from the covariance matrix, equivalent to the determinant when the off-diagonal elements are zero. The same is true for $\boldsymbol{\Sigma_{\hat{y}}}^{-1}$ where the inverse of a diagonal matrix is simply the inverse of each diagonal element. The vectorized closed-form expression of the ELBO is shown in Equation (\ref{eq:ELBO_closed_vec}) where $\boldsymbol{|\Omega|}$ denotes the cardinality of the set $\boldsymbol{\Omega}$, i.e., the number of weight parameters and $N$ denotes the number of classes or output nodes of the final layer.

\section{Continual Learning Methodology}
\subsection{Continual Learning}
Continual Learning aims to retain performance over multiple training periods. Many approaches have been developed to combat the problem of catastrophic forgetting, separating into three main categories: Regularization, Replay, and Dynamic Architectures \cite{Parisi2019}. The Dynamic Architectures approach focuses on adding additional neural resources to a network to account for information in new tasks. Replay focuses on preserving certain data samples from previous tasks and "replaying" them during the training of subsequent tasks. Regularization-based approaches constrain the optimization problem to prevent representations of previous data from adapting exclusively to the new data in the network. The method outlined in this paper aligns with Regularization based Continual Learning, where the KL Divergence regularizes the optimization to restrict parameters to remain close to the previous task.

\subsection{Prior Compression}
Continually learning with Variational Density Propagation leverages the KL Divergence between the variational posterior and the prior in the ELBO objective function. This regulates learning to prevent network parameters from drifting far from the prior. The model parameters are treated as independent random variables to obtain a simplified version of the KL divergence between the variational posterior and the prior, as shown in Equation \ref{eq:ELBO_closed_vec}. This formulation allows each parameter to receive independent gradient updates from the KL term so the new posterior remains close to the previous posterior while receiving updates from the model likelihood. This is an ancillary benefit of assuming all explicitly parameterized random variables are independent. 

When learning over multiple tasks, the prior $p_t(\boldsymbol{\Omega})$ iteratively becomes the posterior of the previous task $q(\boldsymbol{\Omega_{t-1}})$ for task $t$. The updated prior, $q(\boldsymbol{\Omega_{t-1}})$, contains all information from all previously trained task representations, such that $p_t(\boldsymbol{\Omega}) = q_t(\boldsymbol{\Omega_{t-1}})$. For the first task, $t=0$, the prior is chosen as standard normal Gaussian for every network parameter. Using a standard normal prior is known to be parameter sparsity-inducing \cite{Kingma2013}. Using the KL divergence term to constrain network parameters near zero will effectively remove unneeded model parameters from contributing to a model's prediction.This process aligns with the idea of the Minimum Description Length principle \cite{Hinton1993} in which the best model for a given dataset is the one that results in the minimum total description length of the dataset together with the model, satisfying Occam's Razor principle \cite{He2020}. By converging to the minimal model complexity on the first task, all subsequent tasks will retain the goal of minimal model complexity via updating the prior distribution to the variation posterior. Thus, a sparsity-inducing prior is effectively applied across all tasks. This process is shown in Algorithm \ref{alg:CL}.

\begin{algorithm}[H]
\caption{Continual Learning via Prior Compression}\label{alg:CL}
\begin{algorithmic}
\Require $\mathcal{D}=(\mathbf{X},\mathbf{y})$;\\
The predictive distribution $\boldsymbol{\hat{y}}_m\sim\mathcal{N}(\boldsymbol{\mu}_{\hat{y}_m},\boldsymbol{\sigma}_{\hat{y}_m}^2)$; \\
The variational distribution $q_{\boldsymbol{\theta}}(\boldsymbol{\Omega}) \sim \footnotesize{\prod} \mathcal{N}(\boldsymbol{\mu}_{w_j},\boldsymbol{\sigma}_{w_j}^2)$; \\
The prior distribution $p_{0}(\boldsymbol{\Omega}_{0}) \sim \footnotesize{\prod} \mathcal{N}(0,1)$; \\
KL divergence weighting factor $\tau$;
\State $\tau = \tau_0$
\State $q_{\boldsymbol{\theta}_{0}^{*}}(\boldsymbol{\Omega}_{0}) \gets \text{argmax}_{\boldsymbol{\theta}} [\mathop{\mathbb{E}_{q_{\boldsymbol{\theta}}(\boldsymbol{\Omega}_{0})}}[ \ln{p(\boldsymbol{y}|\boldsymbol{X},\boldsymbol{\Omega}_{0})}] -$ \\ 
$\quad\quad\quad\quad\quad\quad\quad\quad\quad\quad\quad\quad\tau\text{KL}_{q_{\boldsymbol{\theta}}(\boldsymbol{\Omega}_{0})}[q_{\boldsymbol{\theta}}(\boldsymbol{\Omega}_{0})||p_0(\boldsymbol{\Omega}_0)]]$ 
\State $\tau = \tau_1$
\For{Task $t > 0$} 
    \State $p_t(\boldsymbol{\Omega}_t) \gets q_{\boldsymbol{\theta}^{*}_{t-1}}(\boldsymbol{\Omega}_{t-1})$ 
    \State $q_{\boldsymbol{\theta}_{t}^{*}}(\boldsymbol{\Omega}_{t}) \gets \text{argmax}_{\boldsymbol{\theta}}[\mathop{\mathbb{E}_{q_{\boldsymbol{\theta}}(\mathbf{\Omega}_{t})}}[ \ln{p(\boldsymbol{y}|\boldsymbol{X},\mathbf{\Omega}_{t})}] - $\\ $\quad\quad\quad\quad\quad\quad\quad\quad\quad\quad\quad\quad\tau\text{KL}_{q_{\boldsymbol{\theta}}(\mathbf{\Omega_{t}})}[q_{\boldsymbol{\theta}}(\mathbf{\Omega}_{t})||p_t(\boldsymbol{\Omega}_t)]]$ 
\EndFor
\end{algorithmic}
\end{algorithm} 

\subsection{Experimental Setup}
Our Continual Learning framework is evaluated for task incremental learning in which task information is given at test time and the network has separate output layers per task, referred to as a multi-headed network \cite{Nguyen2017}. In this multi-headed architecture, all parameters in layers before the classification layer are shared amongst all tasks. Multi-headed networks promote feature sharing by encouraging the slight differences required between tasks to be captured in the bespoke classification layer \cite{Bakker2003}. We evaluate task incremental learning on the MNIST and CIFAR10 benchmark datasets with two versions of task groupings: 5-Split, where there are five different tasks of differentiating between two classes, and  2-Split, where there are two different tasks of differentiating between five classes. Our methodology is also evaluated on Permuted MNIST, in which the MNIST dataset digit's pixels are permuted with ten different functions, one for each task. The network is then tasked with classifying the permuted digits 0-10. For the Split and Permuted MNIST datasets, a fully connected network with a single 1200-node hidden layer is used. For the sequential CIFAR10 datasets, a six-layer convolutional network with three linear layers is used. 
\subsubsection{Hyperparameters}
A hyper-prior was used to initialize the variational posterior where values of mean and variance of the weights were randomly selected from $\mathcal{N}(0, 0.05)$ and $\mathcal{N}(\boldsymbol{\pi}, 0.05)$, respectively. To ensure the positivity in the variance of network parameters, parameter variance values are passed through the soft-plus activation function before actual network operations, shown in Equation \ref{eq:softplus}. For the first task only, the variational prior was set to a sparsity-inducing, the standard normal prior for every parameter, $\mathcal{N}(1,0)$. 

\begin{equation}
    \sigma^2_w = \log(1+\exp(\sigma^2_\pi))
    \label{eq:softplus}\tag{13}
\end{equation}

A grid search is performed to determine the most appropriate initialization scheme for the variance of the weights and KL divergence weighting. The sweep for the mean of the initialization distribution for the variance, $\pi$, considered integer values from -6 to -18. Network layer biases were initialized with the same scheme. The KL divergence term between the variational posterior and the prior manages the trade-off between updates from error embedded in the model likelihood and the divergence from the prior. A weighting term $\boldsymbol{\tau}$ is applied and varied to ensure sufficient representation of each task was learned and retained over multiple tasks. Similar approaches to approximating Bayesian inference in deep learning using Autoencoders \cite{higgins2017} use values of  $\boldsymbol{\tau} > 1$ to promote sparsity in the learned latent representation of the data. Using $\boldsymbol{\tau} > 0.001$ would not allow the network to learn a sufficiently complex network to represent the data when learning the first task with a sparsity-inducing prior. The sweep for the hyperparameter $\boldsymbol{\tau}$ considered values 1e-3 to 1e-6 for the first task and values between 1e-1 and 1e-4 for all subsequent tasks, with increments of powers of ten. Larger values of $\boldsymbol{\tau}$ provide more compression toward the previous tasks posterior. All model variations are trained with the Adam optimizer with an initial learning rate of 1e-3. The learning rate is reduced by a factor of 10 when the loss of each task plateaus. Random seeds are held constant throughout hyperparameter sweeps.

\subsection{Performance Measurement}
After learning all test sets for $t$ tasks, models are evaluated to demonstrate performance and catastrophic forgetting mitigation over multiple tasks. Performance is gauged through the Average Test Classification Accuracy (ACC) and Backward Transfer (BWT). Average Test Classification Accuracy is an average of all test accuracies on individual tasks after training all tasks. Backward Transfer indicates how much learning new information has affected performance on previous tasks. Backward Transfer values less than zero indicate catastrophic forgetting, while values greater than zero indicate improved performance on previous tasks after training on new information \cite{Ebrahimi2019}. If Backward Transfer is greater than zero, then training subsequent tasks improved the generalization of previously trained tasks with information from subsequently trained tasks. These metrics are shown mathematically in Equation \ref{eq:metrics}.
\vspace{-2mm}
\begin{equation}\label{eq:metrics}
    \begin{split}
        \text{BWT} &= \frac{1}{t}\sum^{t}_{i=1}R_{i,t} - R_{i,i} \\
        \text{ACC} &= \frac{1}{t}\sum^{t}_{i=1}R_{i,t}
    \end{split}\tag{14}
\end{equation}

\section{Results and Discussion}
Continual Learning performance with our Variational Density Propagation framework (VDP PC) is compared to Variational Continual Learning (VCL) \cite{Nguyen2017}, which utilizes sampling-based deep variational inference to mitigate catastrophic forgetting via the approximation of Bayesian Inference. Additionally, both approaches are compared to four baselines in which deterministic and VDP frameworks are trained sequentially: a Single-Head deterministic architecture (DET-SH); Fine Tuning (*-FT), where no efforts are made to mitigate catastrophic forgetting; Feature Extraction (*-FE), where shared network parameters are frozen, but bespoke output layers are trained; and Joint Training (*-JT), where later tasks are supplemented with all data from previous tasks. The Average Test Classification Accuracy (ACC) and Backward Transfer (BWT) metrics across both frameworks and all datasets are presented in Table \ref{tab:1}.

\begin{table*}[t]
\normalsize
\caption{Task Incremental Learning Results}
\begin{center}
\begin{tabular}{|c|cc|cc|cc|cc|cc|}
  \hline
  \multirow{2}{*}{\textbf{Approach}} & \multicolumn{2}{c}{\textbf{2-Split MNIST}} & \multicolumn{2}{|c}{\textbf{5-Split MNIST}} & \multicolumn{2}{|c}{\textbf{Permuted MNIST}} & \multicolumn{2}{|c}{\textbf{2-Split CIFAR10}} & \multicolumn{2}{|c|}{\textbf{5-Split CIFAR10}} \\
  \cline{2-11}
  & \textbf{ACC} & \textbf{BWT} & \textbf{ACC} & \textbf{BWT} & \textbf{ACC} & \textbf{BWT} & \textbf{ACC} & \textbf{BWT} & \textbf{ACC} & \textbf{BWT} \\
  \hline
  DET-SH & 50.72\% & -48.42\% & 20.00\% & -79.80\% & 72.81\% & -25.63\% & 46.50\% & -42.89\% & 19.03\% & -74.45\% \\
 \Xhline{4\arrayrulewidth}
  DET-FT & 98.64\% & -0.60\% & 99.34\% & -0.49\% & 96.47\% & -1.67\% & 81.96\% & -18.04\% & 77.68\% & -15.59\% \\
  VDP-FT & 98.55\% & -0.35\% & 98.32\% & -1.46\% & 95.91\% & -2.47\% & 78.82\% &  -8.53\% & 73.42\% & -19.92\% \\
  \hline
  DET-FE & 98.96\% &  0.00\% & 99.33\% &  0.00\% & 96.41\% &  0.00\% & 72.78\% &  -4.24\% & 78.43\% &  -0.79\% \\
  VDP-FE & 97.90\% &  0.00\% & 99.40\% &  0.00\% & 96.72\% &  0.00\% & 79.49\% &   0.02\% & 80.63\% &  -0.13\% \\
  \hline
  VCL    & 98.04\% &  -0.86\% & 99.08\% & -0.50\% & 88.80\% & -7.90\% & 64.39\% & -10.93\% & 76.82\% &  -8.96\% \\
  \textbf{VDP PC} & \textbf{99.24\%} & \textbf{-0.01\%} & \textbf{99.80\%} & \textbf{-0.06\%} & \textbf{97.71\%} & \textbf{-0.14\%} & \textbf{88.81\%} &  \textbf{-1.08\%} & \textbf{83.23\%} &  \textbf{-0.62\%} \\
 \Xhline{4\arrayrulewidth}
  DET-JT & 99.38\% & -0.01\% & 99.87\% & 0.00\% & 98.33\% & -0.10\% & 86.51\% & +0.63\% & 93.11\% & +0.16\% \\
  VDP-JT & 99.34\% & +0.18\% & 99.73\% & -0.08\% & 98.15\% & -0.17\% & 87.81\% & +1.84\% & 94.04\% & +0.19\% \\
  \hline
\end{tabular}
\label{tab:1}
\end{center}
\end{table*}

Despite benchmark datasets' simplistic nature, tasks learned sequentially in a single-headed network (DET-SH) result in a complete loss in performance on previous tasks. Multi-headed networks trained sequentially to fine-tune (*-FT) each new task fare significantly better, retaining the majority of performance after training a sequence of tasks. The bespoke output layer, however, carries the majority of the performance improvement, as demonstrated by freezing all shared parameters (*-FE) before training subsequent tasks and only learning the bespoke output layer. Restricting changes in model complexity with our VDP Prior Compression (VDP PC) approach improves Average Test Classification Accuracy and Backward Transfer performance over all tested sequences of tasks. Performance with the VDP Prior Compression approach closely follows the joint training (*-JT) performance for both the deterministic and VDP frameworks and is considered the upper bound on performance for the tested model architecture on each experiment.

Conversely, we show that Variational Continual Learning (VCL), not trained with a core set, does not improve over existing deterministic baseline approaches. Results collected for VCL are slightly improved over what is reported for Split and Permuted MNIST datasets in the original publication. This improvement is expected to result from reducing the depth of the fully-connected network. Less shared parameters results in less interference from one task to the next. Split Cifar10 results were not collected in the original VCL paper and are demonstrated to follow the same general trend as the MNIST results, performing slightly under fine-tuning performance, demonstrating no additional benefit to catastrophic forgetting mitigation. 

\vspace{-1mm}
\section{Conclusion}
In this work, the Variational Continual Learning framework is improved by removing the requirement for Monte Carlo sampling of the variational posterior of the model parameters. Network operations are replaced with operations of random variables providing a quicker and less noisy estimate of model parameters utilizing a completely closed-form Evidence Lower Bound objective. The inherent compression within the ELBO approximates the Minimum Description Length Principle by penalizing additional model complexity over multiple tasks. Our approach demonstrates catastrophic forgetting mitigation in the task incremental learning setting for a multi-headed network on common Continual Learning benchmark datasets and demonstrates improvement over sampling-based approaches. We leverage the KL regularization weighting term to control the amount of change in model complexity induced by each task. Overall, improved Average Test Classification Accuracy and Backward Transfer metrics are achieved. 

\bibliographystyle{ieeetr}
\bibliography{refs}

\end{document}